\documentclass{svproc}
\usepackage{url}

\usepackage{graphicx}
\usepackage{cite}

\usepackage{xcolor}

\usepackage{caption}
\usepackage{subfigure}
\usepackage{algorithm}
\usepackage{algpseudocode}
\usepackage{amsmath}
\usepackage{amssymb}
\usepackage{pgfplots}
\usepackage{commath}

\definecolor{Yellow}{rgb}{1,0.9,0.7}
\definecolor{Pink}{rgb}{1,0.85,0.85}
\definecolor{AntiqueWhite}{rgb}{0.9,0.9,0.9}

\newcommand{\NOTE}[1]%
{
\noindent
\fboxsep=2mm\fcolorbox{black}{AntiqueWhite}{\parbox{0.95\columnwidth}
{\textbf{NOTE: } #1}
}
}

\algnewcommand\algorithmicswitch{\textbf{switch}}
\algnewcommand\algorithmiccase{\textbf{case}}
\algnewcommand\algorithmicassert{\texttt{assert}}
\algnewcommand\Assert[1]{\State \algorithmicassert(#1)}%
\algdef{SE}[SWITCH]{Switch}{EndSwitch}[1]{\algorithmicswitch\ #1\ \algorithmicdo}{\algorithmicend\ \algorithmicswitch}%
\algdef{SE}[CASE]{Case}{EndCase}[1]{\algorithmiccase\ #1}{\algorithmicend\ \algorithmiccase}%
\algtext*{EndCase}%

\begin{document}
\mainmatter              
\title{Cellular Formation Maintenance and Collision Avoidance Using Centroid-Based Point Set Registration in a Swarm of Drones}
\titlerunning{Cellular Formation Maintenance and Collision Avoidance Using CPSR}  
%

\author{Jawad N. Yasin\inst{1} \and Huma Mahboob\inst{2} \and
Mohammad-Hashem Haghbayan\inst{1} \and Muhammad Mehboob Yasin\inst{3} \and
Juha Plosila\inst{1}}

\authorrunning{J. N. Yasin et al.} 
%
%

\institute{Autonomous Systems Laboratory, Department of Future Technologies, University of Turku, Vesilinnantie 5, 20500 Turku, Finland\\
\email{\{janaya, mohhag, juplos\}@utu.fi},\\
\and
Connected Shopping Ltd, Thetford, UK\\
\email{huma.mahboob@coursemerchant.com}
\and
Department of Computer Networks, College of Computer Sciences \& Information Technology, King Faisal University, Hofuf, Saudi Arabia\\ \email{mmyasin@kfu.edu.sa}}

\maketitle              

\begin{abstract}
This work focuses on low-energy collision avoidance and formation maintenance in autonomous swarms of drones. Here, the two main problems are: 1) how to avoid collisions by temporarily breaking the formation, i.e., \textit{collision avoidance reformation}, and 2) how do such reformation while minimizing the deviation resulting in minimization of the overall time and energy consumption of the drones. To address the first question, we use cellular automata based technique to find an efficient formation that avoids the obstacle while minimizing the time and energy. Concerning the second question, a near-optimal reformation of the swarm after successful collision avoidance is achieved by applying a temperature function reduction technique, originally used in the point set registration process. The goal of the reformation process is to remove the disturbance while minimizing the overall time it takes for the swarm to reach the destination and consequently reducing the energy consumption required by this operation. To measure the degree of formation disturbance due to collision avoidance, deviation of the centroid of the swarm formation is used, inspired by the concept of the center of mass in classical mechanics. Experimental results show the efficiency of the proposed technique, in terms of performance and energy.
\end{abstract}
\section{Introduction}
A swarm is a concept that seems to have no precise definition in literature as such; instead, we find a lot of definitions and discussion addressing swarming i.e. swarm behaviour \cite{Hamann2018}. Swarm robotics can be classified as the study of how a system, consisting of large number of a relatively simple agents, can be designed to attain a desired cumulative behaviour based on the interactions between the agents themselves and between the agents and the environment \cite{jdficc, dorigo2004swarm}. Due to their ability to work in a collaborative style, swarms of autonomous agents add significant advantages over the use of single agent, and therefore they have high demand in diverse fields such as search and rescue, surveying and mapping, inspection, and delivery, in both military and civilian/commercial contexts\cite{jdpaams}. Consequently, the interest of the research community is increasing towards optimization of various autonomy characteristics of drone swarms, for instance collision avoidance, formation maintenance, resource allocation, and navigation \cite{8500274, HE2018327}. In swarm navigation and formation flight, collision avoidance and maintenance of the formation are the most important problems \cite{7935230, jdsurvey}. Formation control methodologies can be categorized into three main approaches: 1) the leader-follower based approach, where every node/drone works autonomously and individually by maintaining a given formation as perfectly as possible by adjusting its position with respect to its neighbours and the leader drone \cite{jy0919, OH2015424, jdpaams}; 2) the behaviour based approach, where, based on a pre-determined strategy, one behaviour is chosen out of the available ones \cite{1261347, 736776}; and 3) the virtual structure based approach, where the swarm is considered a single entity, i.e. a single large drone effectively, and navigated through a trajectory accordingly \cite{4586750, DONG2016415}.

Cellular automata based modelling provides the environment that each cell can decide its movement by only looking at its neighbours and the environment and based on its rules that are defined for each individual cell dynamically and run-time \cite{noauthor_stephen_nodate}. The modelling of cellular automata in our collision avoidance algorithm provide us the opportunity to reform the system to pass the obstacle only by defining some distributed rules for each individual drone. In other words to pass the collision we do not need to have a central processing element that defines the path of each drone. In return each node, individually, in a dynamic and flexible movement can pass the obstacle so that the overall time and energy consumption of the swarm is optimized. To define these rules for individual drones, we make use of genetic algorithm techniques that is highly compatible with cellular automata model.

A genetic algorithm (GA) is one of the simplest random-based classical evolutionary computing methods, where random changes are applied to the current solutions to generate new solutions for finding an optimal or near-optimal solution \cite{gad_introduction_2018}. GAs work by utilizing the basic principles of generation of potential random solutions, selection of the best solution by calculating the distance of each solution to the destination, generation of new solutions based on the generated good solutions, and repeating these steps in order to reach the desired result \cite{174685, 6942974}. The ability of GAs to converge close to the global optimum and their relatively simple implementation make them quite popular among available optimization heuristics \cite{bansal2019evolutionary}.

Point set registration is a commonly used method, playing an important role in various applications such as image retrieval, 3D reconstruction, shape and object recognition, and SLAM \cite{8594514}. In point set registration, the correlation between two point sets is determined in order to retrieve the required transformation that maps one point set to the other \cite{DBLP:journals/corr/abs-0905-2635}.

Our algorithm has two phases: 

\begin{itemize}
    \item The first phase is a cellular automata based collision avoidance scheme that disturbs the original formation to pass the obstacle.
    \item The second phase is a re-formation scheme, inspired by point set registration, that will resume from the highest disturbed formation to the original formation.
\end{itemize}

In this paper, the leader-follower based approach is utilized for drone swarm control due to its reliability, ease of implementation and analysis, and scalability  \cite{jdsensors, 10.1007/978-3-319-18944-4_10}. However, in our proposed solution, there is no unique global leader, as the leader gets changed dynamically based on certain constraints. The cellular formation and collision avoidance algorithms are integrated with a simple GA-inspired approach and a point set registration method in order to optimize the collision avoidance and re-formation phases. The goal is to calculate the escape routes and select a near-optimal path upon detection of an obstacle, having minimal deviation from the original route. Once a defined danger zone has been passed, for reconstruction of the formation, centroid-based point set registration (CPSR) is used in the formation maintenance algorithm to optimally bring back each drone that lost its position in the formation when avoiding a collision with the detected moving obstacle.

Using a GA-inspired approach for collision avoidance in a swarm of robots/UAVs is beneficial as it allows the algorithm to check for all possible maneuvers and select the best solution depending on the pre-defined constraints such as the minimal movement requirement and power consumption limitations. Furthermore, with the help of CPSR, once a collision has been avoided, the formation can be obtained again swiftly and optimally by bringing the UAVs back to the desired formation shape. CPSR facilitates this dynamic recovery process and autonomous switching of the swarm leader according to the requirements posed by the scenario at hand. This can be very helpful especially in cases where the time to complete a mission is critical.

The rest of the paper is structured as follows. In Section 2, the proposed algorithm is described. Section 3 provides the simulation results. Finally, Section 4 concludes the paper with some discussion and comments on future work.

\section{PROPOSED APPROACH}

The general pseudo code of the proposed approach is given in Algorithm \ref{algo1}. We start with the assumption that the leader-follower connection has been established and that the formation is already maintained before a mission starts. By utilizing the on-board processing units, this top-level algorithm is executed by every individual node locally. Algorithm \ref{algo1} starts by initializing the Boolean variable/flag $FLAG_{obs}$ (Line 2) whose role is to indicate absence ($False$) or presence ($True$) of an obstacle.
Then the target shape (\textit{TShape}) of the swarm is initialized based on the current state or position of each node with respect to the others (Line 3). \textit{TShape} is the next targeted formation shape calculated at every interval for the next time interval, determining the next target position for each node to propagate to. 

After the above initial steps, the main loop (Lines 4-13) is entered. First, the procedure \textit{ObstacleDetection} is called, and based on the values of the variables/signals calculated by this algorithm, information on the presence and characteristics of a potential obstacle is available (Line 5). In case an obstacle is detected, i.e., $FLAG_{obs}$ $==$ $True$, the procedure \textit{CollisionAvoidance} is called, and \textit{TShape} is set up according to the feedback received (Lines 6-7). After this, once a collision has been successfully avoided, $FLAG_{obs}$ is reset to $False$ (Line 8). On the other hand, if no obstacle is detected, i.e., if $FLAG_{obs}$ $==$ $False$ holds after the execution of \textit{ObstacleDetection}, then \textit{TShape} is updated without an involvement of the collision avoidance procedure (Line 10).  Finally, the point set registration based re-formation algorithm \textit{PS-ReFormation} is called to re-establish the desired formation (Line 12). This has an effect only if the formation is distorted due to a collision avoidance event. 

It is important to note here that in the point set registration of the re-formation process, it is crucial to optimally and rapidly calculate the mapping between the current and desired shapes of the swarm. This is the case especially when complicated movements are involved which drastically change \textit{TShape} of the swarm, for instance, when the angle of the leader's movement strongly changes due to the presence of an obstacle on the path. 

\begin{algorithm}
\caption{Global Routine}\label{algo1}
\begin{algorithmic}[2]
\Procedure{Obstacle Detection \& Navigation}{}
\State{$FLAG_{obs}$ $\gets$ $False$;}
\State{TShape $\gets$ Initialization based on current state;}
\While{True}
    \State{$FLAG_{obs}, D_{obs}, A_{obs}, V_{obs}, D_{impact}$ $\gets$ Obstacle\par
        \hskip\algorithmicindent Detection();}
    \If{$FLAG_{obs}$}
        \State{TShape $\gets$ CollisionAvoidance($D_{obs}, A_{obs}, V_{obs}, D_{impact}$);}
        \State{$FLAG_{obs}$ $\gets$ $False$;}
    \Else
        \State{Update TShape;}
    \EndIf
    \State{PS-Reformation(TShape);}    
\EndWhile
\EndProcedure
\end{algorithmic}
\end{algorithm}

\subsection{Obstacle detection}
\begin{figure}[!ht]
    \centering
    \includegraphics[width=0.5\textwidth]{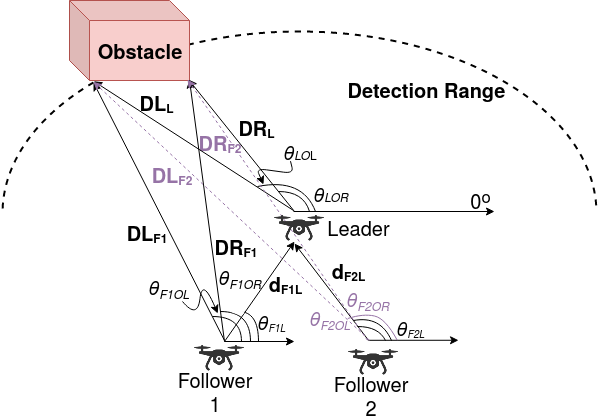}
    \caption{Distance and Direction Calculation \label{fig:coordcalc}}
\end{figure}

\begin{figure}[!ht]
    \centering
    \includegraphics[width=0.5\textwidth]{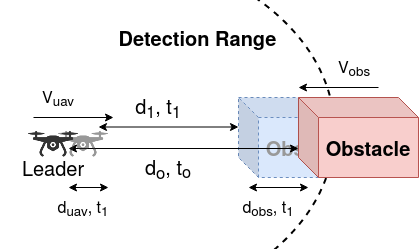}
    \caption{Moving Obstacle Calculation \label{fig:movobs}}
\end{figure}

\begin{algorithm}
\caption{Obstacle Detection}\label{algo2}
\begin{algorithmic}[2]
\Procedure{Obstacle Detection()}{}

\If{$obstacle$ in $Detection\_Range$}
    \State{$FLAG_{obs}$ $\gets$ $True$;}
    \State {$D_{obs}$, $A_{obs}$ $\gets$ Calculate obstacle distance and angles at which\par
    \hskip\algorithmicindent the edges lie;}
    \State{$V_{obs}$ $\gets$ Calculate obstacle Velocity;}
    \State{$D_{impact}$ $\gets$ Calculate distance to impact;}
    \State{return($FLAG_{obs}, D_{obs},A_{obs}, V_{obs}, D_{impact}$)}
\EndIf
\EndProcedure
\end{algorithmic}
\end{algorithm}
In this procedure (specified in Algorithm \ref{algo2}), the node continuously scans for the obstacles, and soon as there is an obstacle in the detection range of the on-board sensor system, the signal flag $FLAG_{obs}$ is set to $True$ (Lines 2-3). After this, the calculation of the parameters of the obstacle is done, i.e., the distance to the obstacle ($D_{obs}$) and the angle at which the detected obstacle lies $A_{obs}$ (Line 4), as shown in Figure \ref{fig:coordcalc}. Then it is determined if the obstacle is moving or stationary (Line 5), using Equations (1)-(3) and illustrated in Figure \ref{fig:movobs}. The velocity of the obstacle ($V_{obs}$) is computed, and based on the value of $V_{obs}$, we have three possible case scenarios: (1) if \textbf{$V_{obs}$ == 0} then the environment is static or the obstacle under observation is stationary; (2) if $V_{obs}$ is negative, then the obstacle is coming towards the UAV; or (3) if $V_{obs}$ is positive, then the obstacle is going away from the UAV. Based on the computed velocity of the obstacle, distance to the potential impact ($D_{impact}$) is calculated (Lines 5-6). These calculations are elaborated in the equations below.

\begin{table}[H]
    \caption{Description of Variables from Fig. \ref{fig:coordcalc}}
    \centering
    \scriptsize
    \begin{tabular}[b]{|c|l|}
    \hline
    Variables & Description \\
    \hline
    $DR_i$ & distance of right and left edges\\
    $DL_i$ & of the obstacle from leader, follower 1 \\
    & and follower 2 as shown in Figure\\
    \hline
    d\textsubscript{F1L} & distance of leader from follower 1 and\\
    d\textsubscript{F2L} & follower 2 respectively\\
    \hline
    $\theta\textsubscript{LOR}$ & angle at which right and left edges are\\ $\theta\textsubscript{LOL}$ & detected from leader respectively\\
    \hline
    $\theta\textsubscript{F1L}$ & angle of leader from follower 1 and\\
    $\theta\textsubscript{F2L}$ &  follower 2, respectively\\
    \hline
    \end{tabular}
    \label{tab:table1}
\centering
\end{table}

We know that after $t_{1}$ seconds, the distance travelled by the UAV can be calculated by:

\begin{equation}
    d_{UAV} = v * (t_{1} - {t_0})
\end{equation}

Then the distance travelled in the meantime by the obstacle and the velocity of the obstacle are calculated by (2) and (3), respectively:

\begin{equation}
    d_{obs} = d_o - d_{UAV} - d_1
\end{equation}
where $d_o, d_1$ are the distances between the UAV and the obstacle at times $t_o$ and $t_1$ respectively.
\begin{equation}
    v_{obs} = d_{obs} / \Delta t 
\end{equation}

\begin{figure}[!ht]
\begin{center}
    \subfigure[\label{fig:ptimdanger}]{\includegraphics[width=0.35\columnwidth]{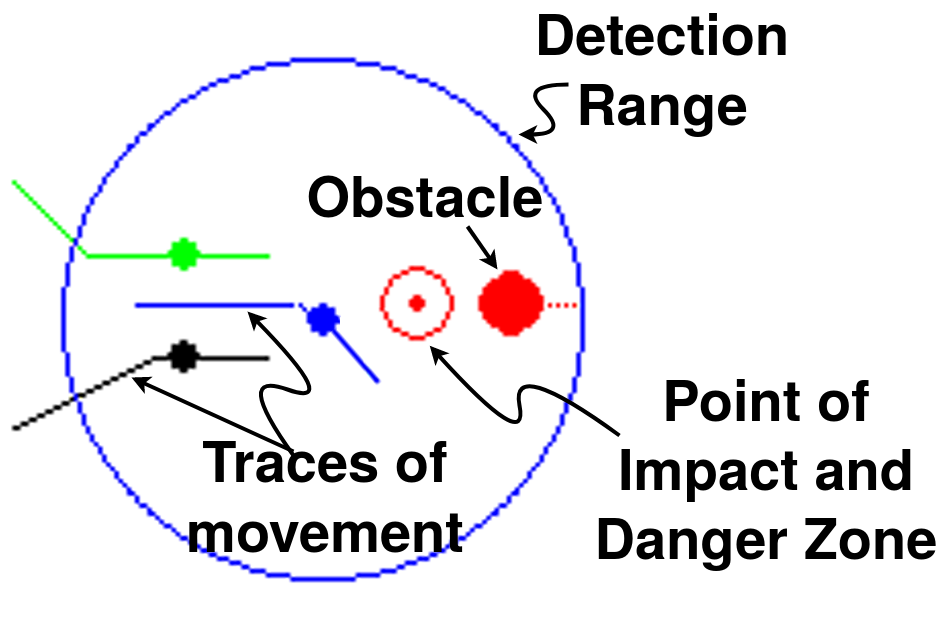}}
    \subfigure[\label{fig:dstrbnce}]{\includegraphics[width=0.35\columnwidth]{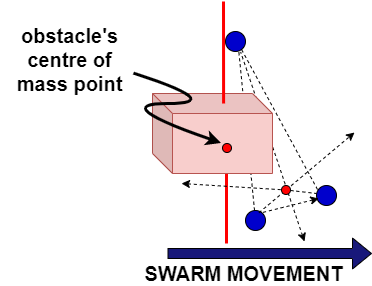}}
   \end{center}
  \caption{(a) Point of impact and Danger Zone as they appear dynamically. (b) Highest level of disturbance illustration}
  \label{fig:bvlahblah}
\end{figure}

In Eq. (2), if \textbf{$d_{obs}$ == 0}, it means the obstacle is stationary. Otherwise the obstacle is moving and in that case if the distance between the obstacle and the UAV after $t_{1}$, i.e., $d_{1}$ is less than the distance detected at time $t_{o}$, i.e., $d_{o}$, then the obstacle is moving towards the UAV (Figure \ref{fig:movobs}). If $d_{1}$ is greater than $d_{o}$, it means the obstacle moving away from the UAV. Figure \ref{fig:ptimdanger} shows the point when the obstacle has entered the detection range of the UAV. The obstacle's trace of movement is shown as a red dotted line; similarly, the UAVs' traces of movement are shown as correspondingly coloured lines behind the smaller coloured circles representing three UAVs. Based on the movement of the obstacle, the computational \textit{point of impact} and the dimensions of the obstacle are calculated, and based on these the Danger Zone (the red circle, the point of impact is the red dot inside it) is defined as shown in Figure \ref{fig:ptimdanger}.

\subsection{Collision Avoidance}

\begin{figure}[!ht]
\begin{center}
    \subfigure[\label{fig:cellu1}]{\includegraphics[width=0.49\columnwidth]{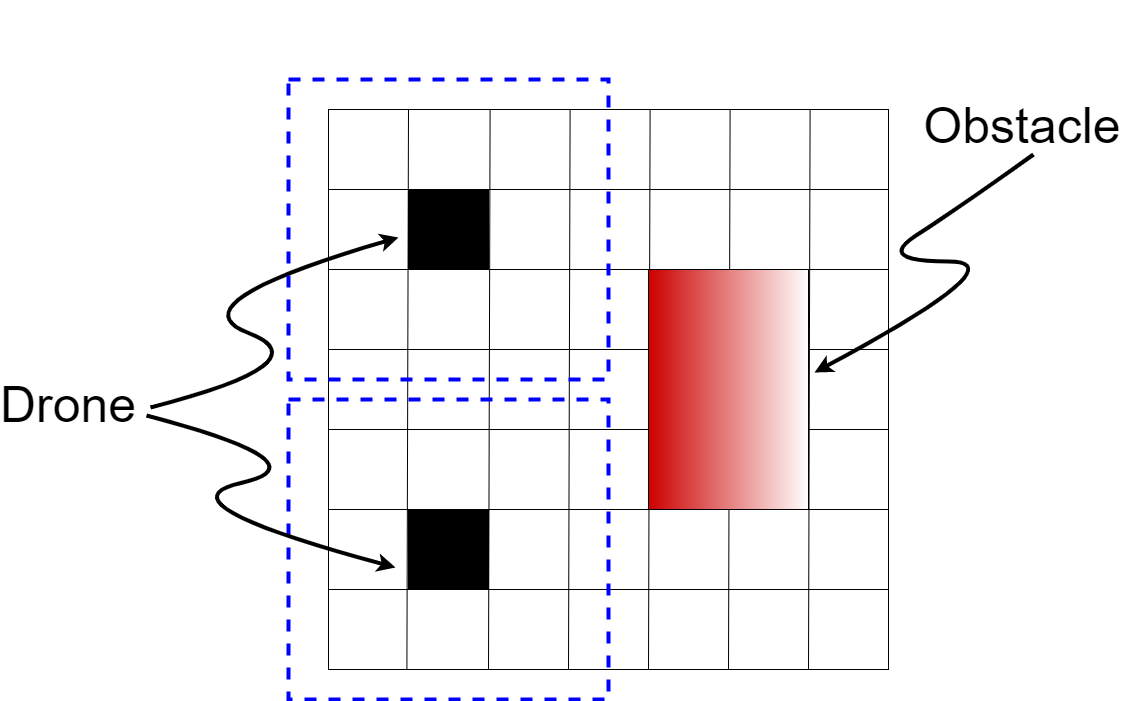}}
    \subfigure[\label{fig:cellu2}]{\includegraphics[width=0.49\columnwidth]{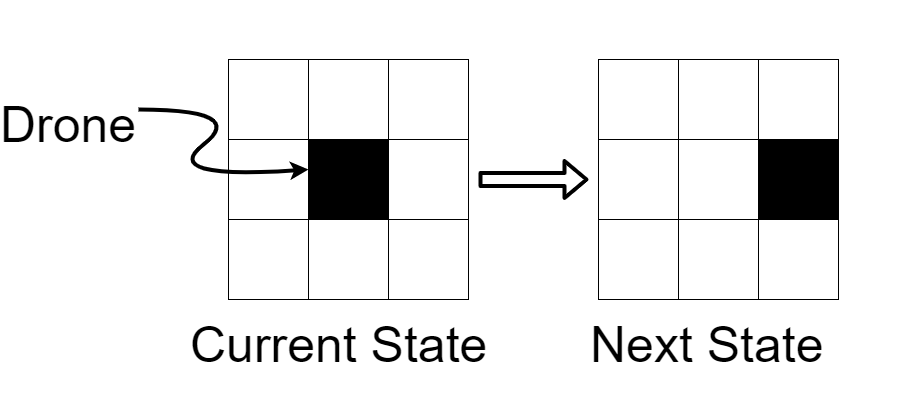}}
    \caption{(a) 2D model of 2 drones and one obstacle. (b) Example of movement according to a rule.}
   \end{center}
\end{figure}

Collision avoidance in our proposed algorithm is simply defining continuously the next step formation for the swarm in a way that this sequence of formations can pass the obstacle. This re-forming of the swarm continues until the swarm reaches the \textit{highest formation disturbance} in which it is guaranteed the swarm can pass the obstacle. The highest formation disturbance is defined as the state when all the drones have passed the line perpendicular to the velocity vector of the swarm, i.e., \textit{swarm movement} in Figure \ref{fig:dstrbnce},  and passing through the mass point of the obstacle, see Figure \ref{fig:dstrbnce}. After that the swarm resumes back to its original formation via TPS-based algorithm \cite{jdtps} that will be discussed in the reformation section\footnote{Even though the highest formation disturbance state might not totally guarantee not to have collision in TPS-based reformation phase, we take this assumption to simplify finding the moment of switching from GA-based collision avoidance phase to TPS-based resuming the original formation.}. In this section we only cover the collision avoidance algorithm until reaching the highest level of disturbance. The goal here is to determine rules for each drone by which the drones pass the obstacle in a way that the time and energy minimizes. To do this we made use of applying GA in a cellular automata (CA) model of the swarm. This model is based on separating the space, 2D or 3D, into identical grid zones where the size of the grid is determined to encompass one drone in its safe distance from the borders of the grid. Using this modeling method, 2D/3D environment can be divided to identical grid zones, \textit{cells}, where existences and non-existence of a drone in each grid can change the state of the cell to one and zero respectively. Figure \ref{fig:cellu1} shows such a model for 2 drones and one obstacle in 2D environment. Each cell in this model only can see its neighbors, like standard CA \cite{noauthor_stephen_nodate}, and based on the neighbors it decides its next state. If the cell is occupied by a drone, the next state of the cell is determined via the movement of the drone that is limited towards cardinal and inter-cardinal directions (Figure \ref{fig:cellu2}). 

Our GA-based algorithm tries to find the best possible rule for each cell occupied by the drone i.e., black, that minimizes the time. The time in this model is the number of steps for the swarm to reach the highest level of disturbance. The energy in each step, i.e., for the CA rule, can only have three values, 1) '0', when the drone stays in its position, 2) '1', when there is a movement toward one cardinal direction, and 3) '2', when there is a movement toward one inter-cardinal direction. The overall energy for a drone is the summation of energy consumption in each iteration from the start of reformation to highest formation disturbance state.

\begin{algorithm}
\caption{CollisionAvoidance}\label{algo3}
\begin{algorithmic}[2]
\Procedure{CollisionAvoidance($D_{obs}, A_{obs},V_{obs}, D_{impact}$)}{}
\State{$DangerZone$ $\gets$ Calculate based on $D_{impact}$ and obstacle dimensions;}
\While{$DangerZone$}
    \State{Calculate Escape routes;}
    \State{Update the TShape using GA;}
\EndWhile
\State{return(TShape)}
\EndProcedure
\end{algorithmic}
\end{algorithm}

The population of the potential solutions was done by applying the following principles:

\begin{itemize}
    \item generation of potential random solutions by defining different rules for each drone
    \item calculate the time and energy of each solution when reaching the highest disturbance formation is targeted, i.e., when all the drones pass the obstacle,
    \item regenerate new solutions by mutation of good obtained rules for each drone
\end{itemize}

These routines are integrated with the collision avoidance developed and presented in \cite{jy0919}, in such a way that the translated \textit{TShape} destination of each node of the swarm is checked at each time interval. If there is an obstacle in the path of a node, the \textit{TShape} destination may not be the same as the original destination. Therefore, the \textit{TShape} destination is calculated by using a fixed grid around the danger zone in order to restrict the GA from populating infinite exhaustive solutions. Afterwards at each iteration, the way to reach the \textit{TShape} destination is optimized by point set registration.

\subsection{Re-Formation}

Observing and avoiding an obstacle by the swarm, in most of the cases changes/disturbs the formation of the swarm until reaching the highest disturbance formation where after the swarm must be restored to the initial formation state. This process raises a formation construction problem that is widely covered in the literature. However, in our case, the re-formation algorithm, or in other words the \textit{disturbance rejection} of a swarm, must be compatible with our obstacle detection and collision avoidance algorithm whose main target is to reduce the overall settling time and energy of the system, i.e., bringing the disturbed centroid back to its intended state in the \textit{TShape}. It is worth mentioning that in the process of resuming the formation it is not necessarily needed to keep the initial neighbouring state among the drones since in the formation all the drones are considered to be an identical node. Furthermore, since there is no dedicated leader and leader is only selected according to the situation at hand, therefore the dynamicity of re-formation process gets smoother with no pauses or unnecessary waiting times for nodes. For example, in the original state if drone 1 has two neighbors drone 2 and 3, after the reconstruction and resuming the formation this might not necessarily happen. Or as shown in next section (Figure \ref{fig:ini_detect}), the leader before the swarm gets disturbed and after the reformation process is not the same, as the leader was re-elected dynamically.

In the process of resuming back from the disturbed state of the swarm formation, referred to as the \textit{scene} in this section, to the initial formation state, i.e., the \textit{TShape model}, two main questions are: 1) what is the optimal alignment or \textit{mapping} of identical \textit{nodes} in the disturbed formation of the swarm, i.e., the scene, and in the initial formation, i.e., \textit{TShape} model; and 2) what is the optimal trajectory of each node in the scene to be \textit{mapped} into the corresponding node in the \textit{TShape} model? For the first problem, we adopt a well-know idea in point set registration \cite{8594514, DBLP:journals/corr/abs-0905-2635} that is based on the thin-plate spline (TPS) technique that is used in data interpolation and smoothing \cite{854733}. After determining the mapping strategy, for the second problem, we use the shortest path scheme when applying the proposed collision avoidance approach. In the following, we first explain the concept of thin-plate splines, and after this we propose an algorithm based on that.

A piece-wise function defined by polynomials is known as a spline. Complex and complicated shapes are approximated with ease via curve fitting using splines due to their non-complicated construction \cite{854733}. For simplicity, we discuss the algorithm for 2-dimensional formulation and presume to have two sets of correlating data sets or points $X$ i.e. {$x_i$, i = 1, 2, . . . , $n$} and $V$ i.e. {$v_i$, i = 1, 2, . . . , $n$}. Where $x_i$ and $v_i$ are the coordinate representation of the locations of a point, $x_i$ = (1, $x_ix$, $x_iy$) and $v_i$ = (1, $v_ix$, $v_iy$), in the scene and model respectively. Considering the shape of the disturbed function, finding a mapping function $f(v_i)$ that fits between correlating point sets $X$ and $V$ can be obtained by minimizing the following:

\begin{equation}
\begin{split}
    E_{TPS}(f) = \sum_{i=1}^{n}||x_i - f(v_i)||^2 + \\ \lambda\iint[(\frac{\partial^2f}{\partial x^2})^2 +2(\frac{\partial^2f}{\partial x\partial y})^2 + (\frac{\partial^2f}{\partial y^2})]dxdy
\end{split}
\label{etps}
\end{equation}

Where $E_{TPS}$ is the energy function that is considered as the measurement for the amount of formation disturbance. The integral part of the equation represents how the corresponding point sets are mapped to the correlating point set by keeping the intended \textit{formation} under consideration. Also, the factor $\lambda$ provides the scaling. If the intention is to map one point set over the other without considering the shape of the disturbed swarm, $\lambda$ should be set to zero and the closest points are mapped accordingly without keeping the shape under consideration. In this situation, the disturbance, i.e., $E_{TPS}$, is simply as follows:

\begin{equation}
\begin{split}
    E_{TPS}(f) = \sum_{i=1}^{n}||x_i - f(v_i)||^2 
\end{split}
\end{equation}

\begin{figure}[!ht]
    \centering
    \includegraphics[width=0.65\textwidth]{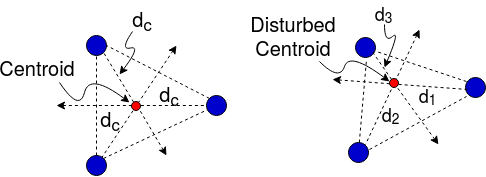}
    \caption{Centroid of the swarm \label{fig:centroid}}
    \label{Fig:centroid}
\end{figure}

Minimization of such a temperature function determines the mapping process from the disturbed swarm in the highest disturbance formation, i.e., the scene, to the original shape after the obstacle. Via this mapping the new leader also will be determined. After calculating the mapping function, each drone from the scene follows the shortest path to reach its hypothetical location in the model. Since the model in Equation \ref{etps} is hypothetical and uncertain events might affect the process of reformation, a relative run-time measurement is needed to continuously assure that the swarm is reaching its formation. This metric is the hypothetical center of the swarm that can be calculated from an instantaneous location of the drones in the swarm. The continuous error in the formation that should be dynamically observed and reduced is calculated by summation of the deviation of the distance of each drone from the center w.r.t. the golden formation model, that is shown by $d_{rms}$. As an example, Figure \ref{fig:centroid} shows the centroid point for three drones in the golden formation model, i.e., the left side, and the disturbed model, i.e., the right side. Based on this, the deviation of the distance of each drone from the center w.r.t. the golden formation model is as follows:

\begin{align}
\label{eqn:centroideq}
\begin{split}
    \Delta d_1 = d_c - d_1\\
    \Delta d_2 = d_c - d_2\\
    \Delta d_3 = d_c - d_3
\end{split}
\end{align}

and

\begin{equation}
    d_{rms} = \sqrt{\Delta d_1^2 + \Delta d_2^2 + \Delta d_3^2}
\end{equation}

The measurment of $d_{rms}$, is a figure of merit. From this equation it is determined how much the current formation has been distorted from its original/pre-defined formation. So, minimizing the $d_{rms}$ to zero will bring back the formation optimally, i.e., $d_{rms}$ $\longrightarrow$ 0.

The reformation process is done by first calculating the centroid of the swarm as shown in the Figure \ref{fig:centroid}.

These values are then fed to the point set registration, in order to calculate the optimal solution for bringing the UAVs back to the desired formation, and in the meantime, bringing the centroid as quickly as possible to the final destination, which can be seen in the next section, i.e, Simulation \& Results. During the re-formation the leader of the swarm changes dynamically, and the previous leader (UAV1/Blue UAV) goes to the position of the current leader (UAV2/Green UAV). The reason for that is, while UAV1 is deviating from its current trajectory in order to avoid colliding with the obstacle, it slows down. In the meantime UAV2, which continues in its path, becomes a more likely candidate for going to the position of UAV1 rather than slowing down for it. Therefore, UAV2 moves to the location of UAV1, and simultaneously UAV1 moves to the previous location of UAV2, as soon as UAV1 has successfully avoided the collision. 

\section{SIMULATION \& RESULTS}

\begin{figure}[!ht]
\begin{center}
    \subfigure[\label{fig:initial1}]{\includegraphics[width=0.4\columnwidth]{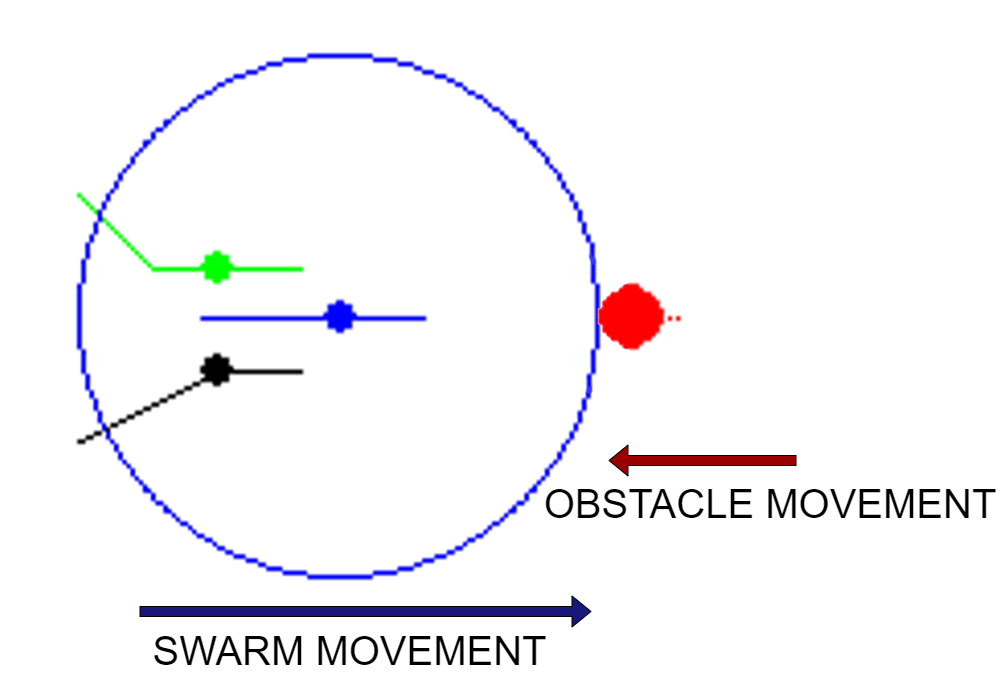}} \hspace{0.2cm}
    \subfigure[\label{fig:obs_detect}]{\includegraphics[width=0.4\columnwidth]{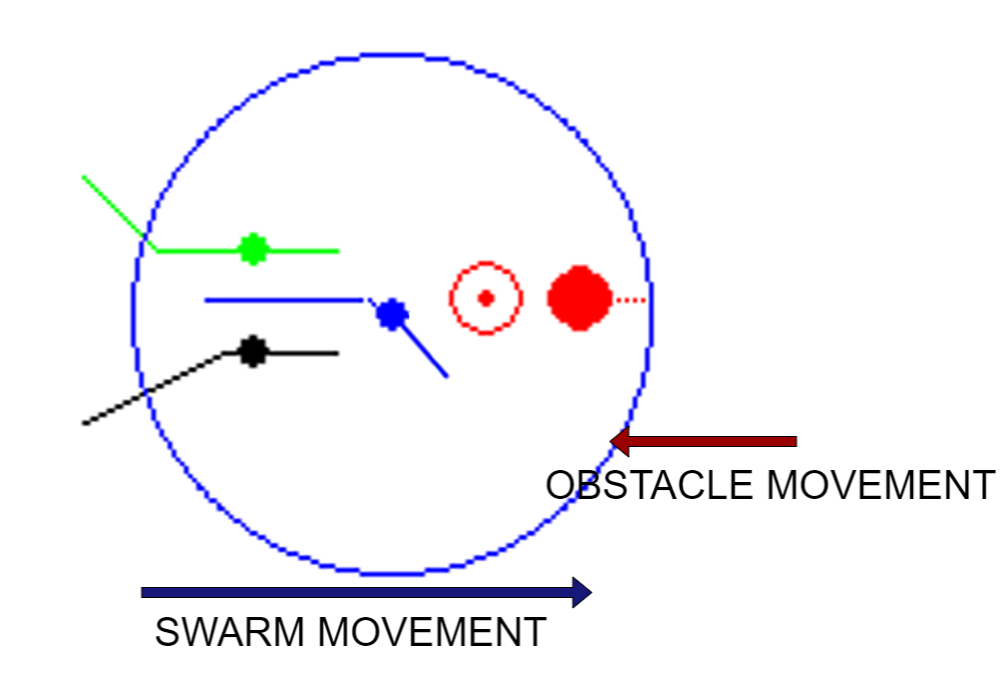}}
    \subfigure[\label{fig:bypasing}]{\includegraphics[width=0.4\columnwidth]{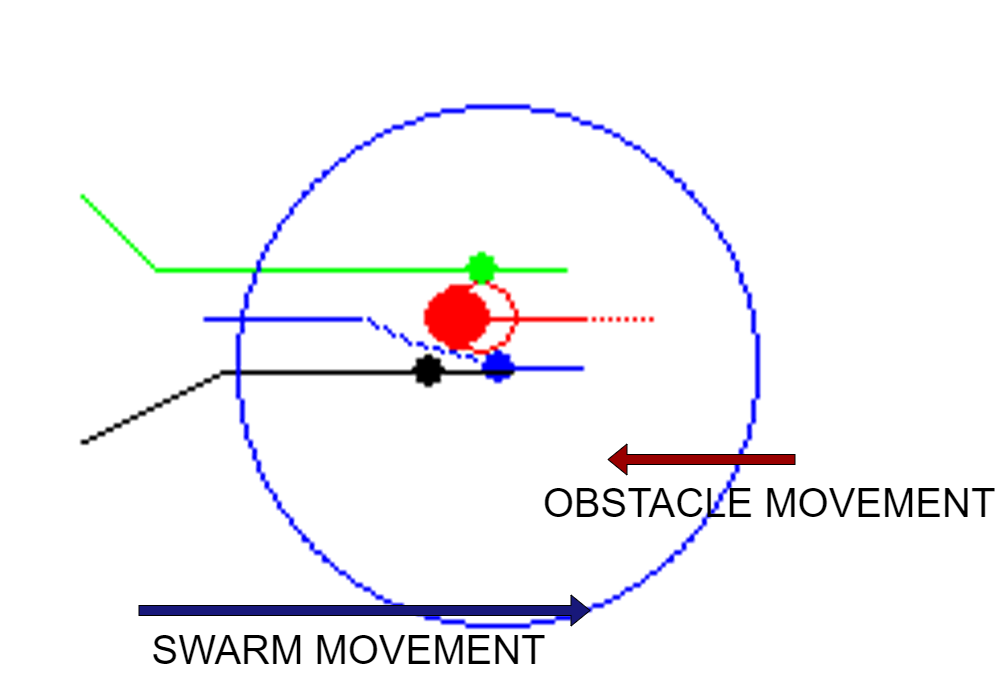}} 
    \subfigure[\label{fig:uav2ahead}]{\includegraphics[width=0.4\columnwidth]{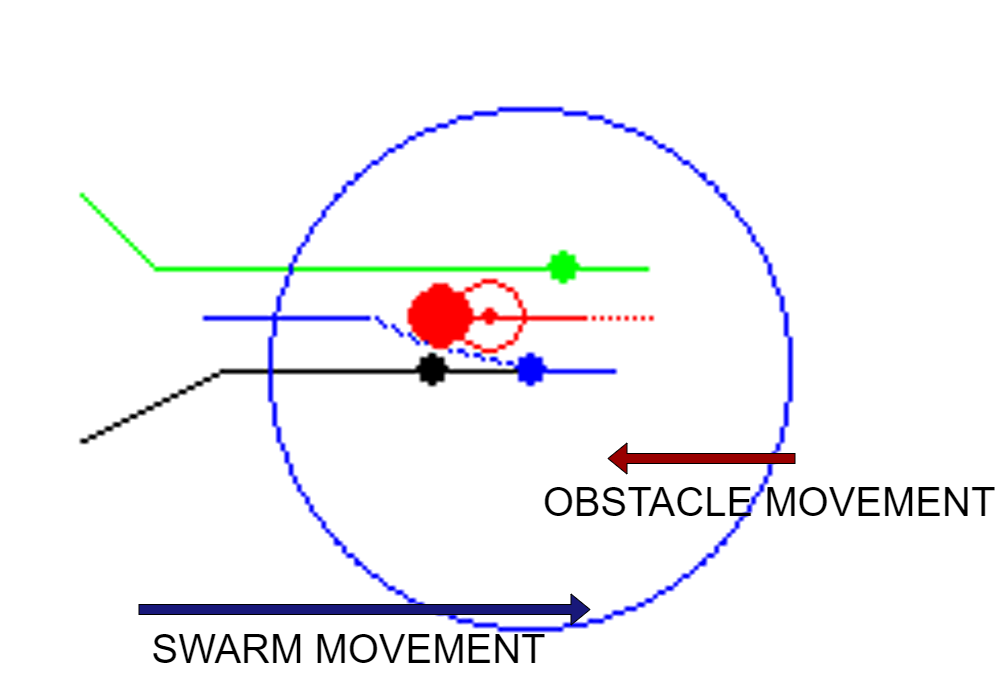}}
   \subfigure[\label{fig:leaderchange}]{\includegraphics[width=0.4\columnwidth]{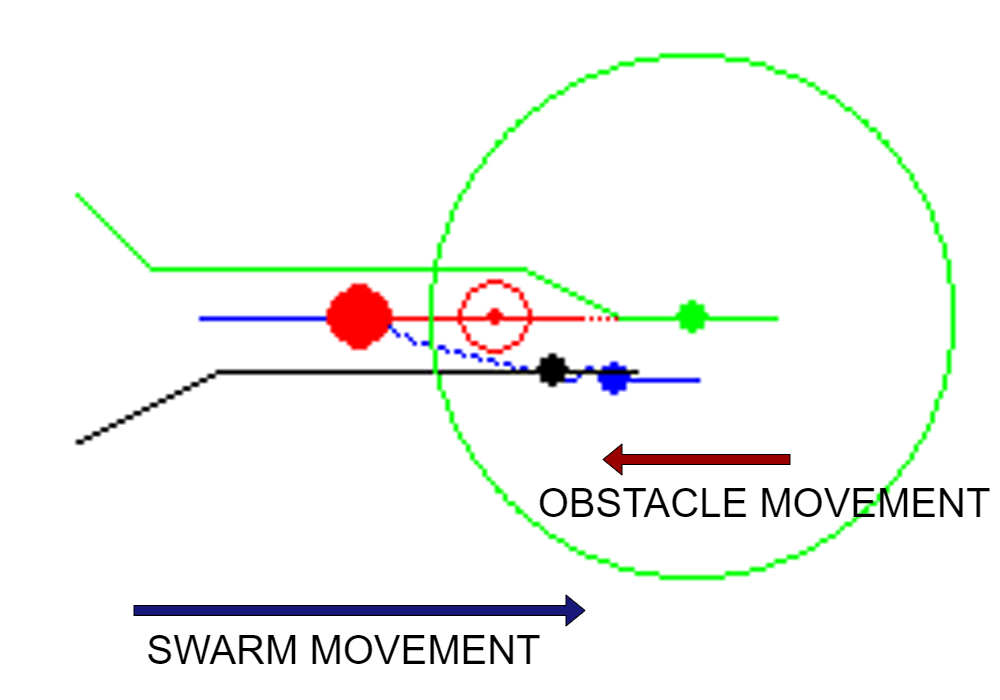}}
   \end{center}
  \caption{Different time intervals from Spawning to when the obstacle comes in detection range. (a) When the UAVs are spawned and obstacle is moving towards the swarm. (b) obstacle is in detection range of the UAV, and Point of Impact is calculated and shown, UAV1 is deviating from its original path in order to avoid the Danger Zone. (c) Bypassing the obstacle. (d) bypassing the obstacle 2. (e) Leader changed while bypassing the obstacle.}
  \label{fig:ini_detect}
\end{figure}

\begin{figure*}[!ht]
\begin{center}
    \subfigure[\label{fig:init1}]{\includegraphics[width=0.35\columnwidth]{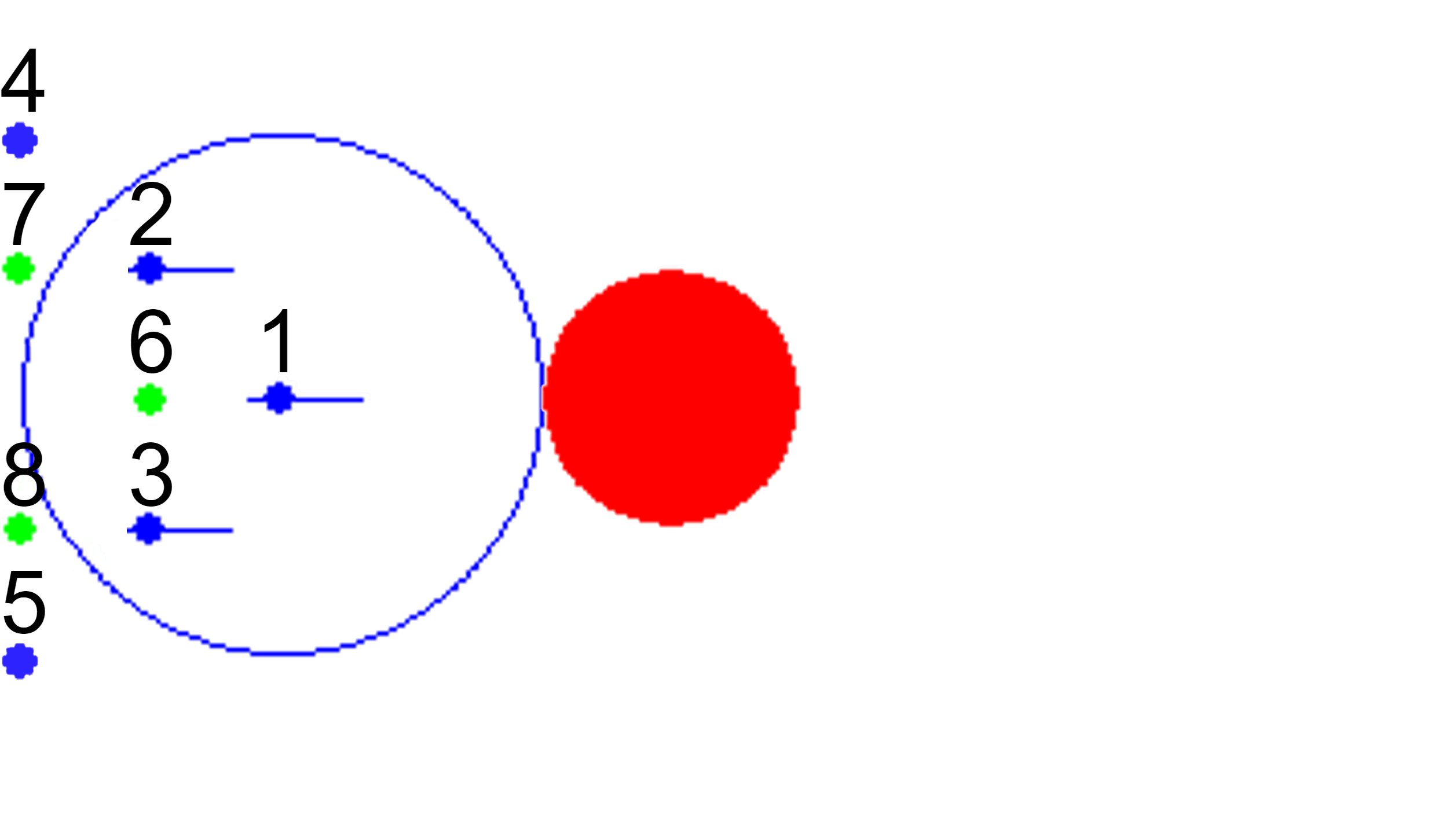}} \hspace{0.2cm}
    \subfigure[\label{fig:init2}]{\includegraphics[width=0.35\columnwidth]{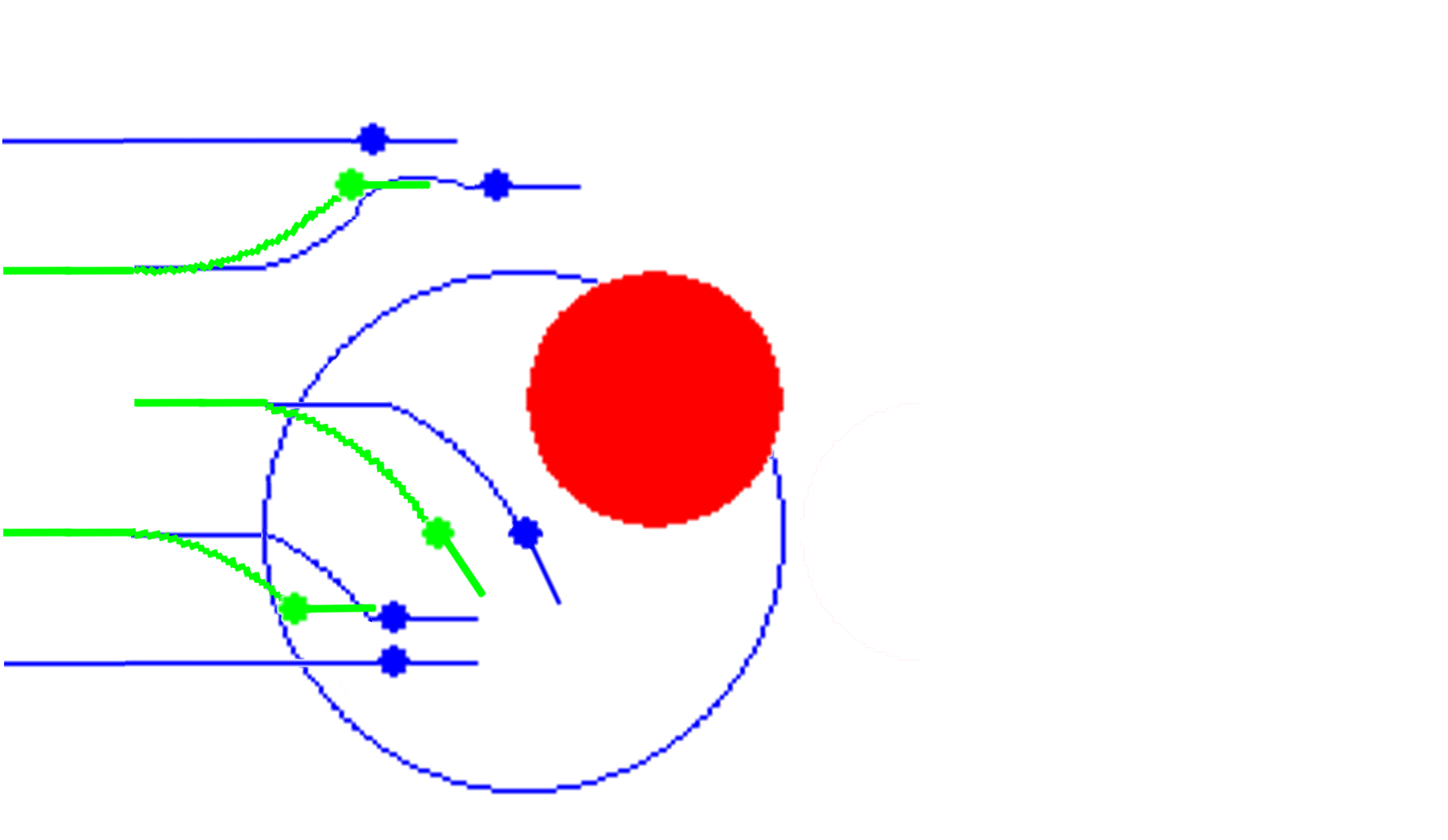}}
    \subfigure[\label{fig:init3}]{\includegraphics[width=0.35\columnwidth]{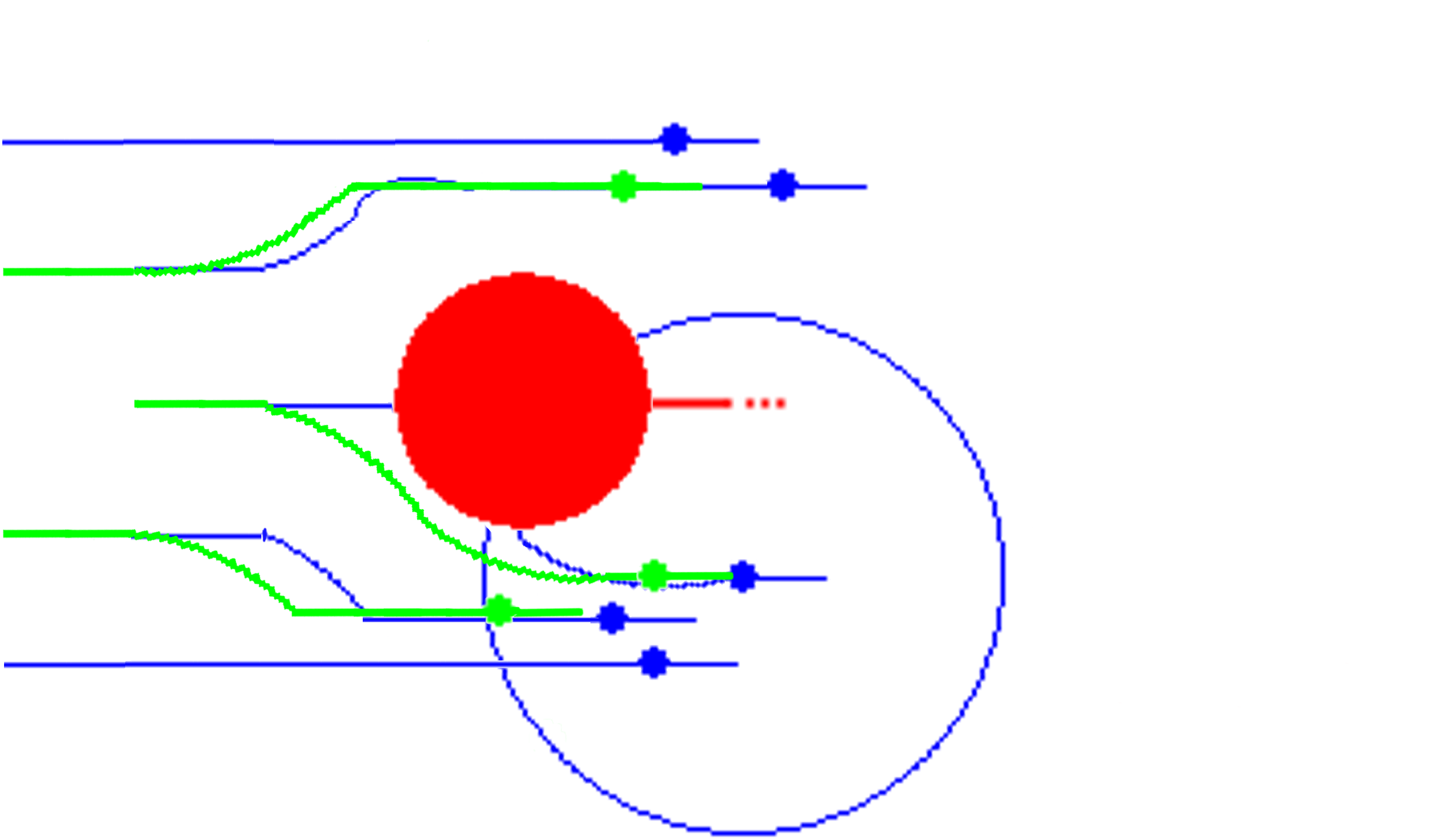}} 
    \subfigure[\label{fig:init4}]{\includegraphics[width=0.35\columnwidth]{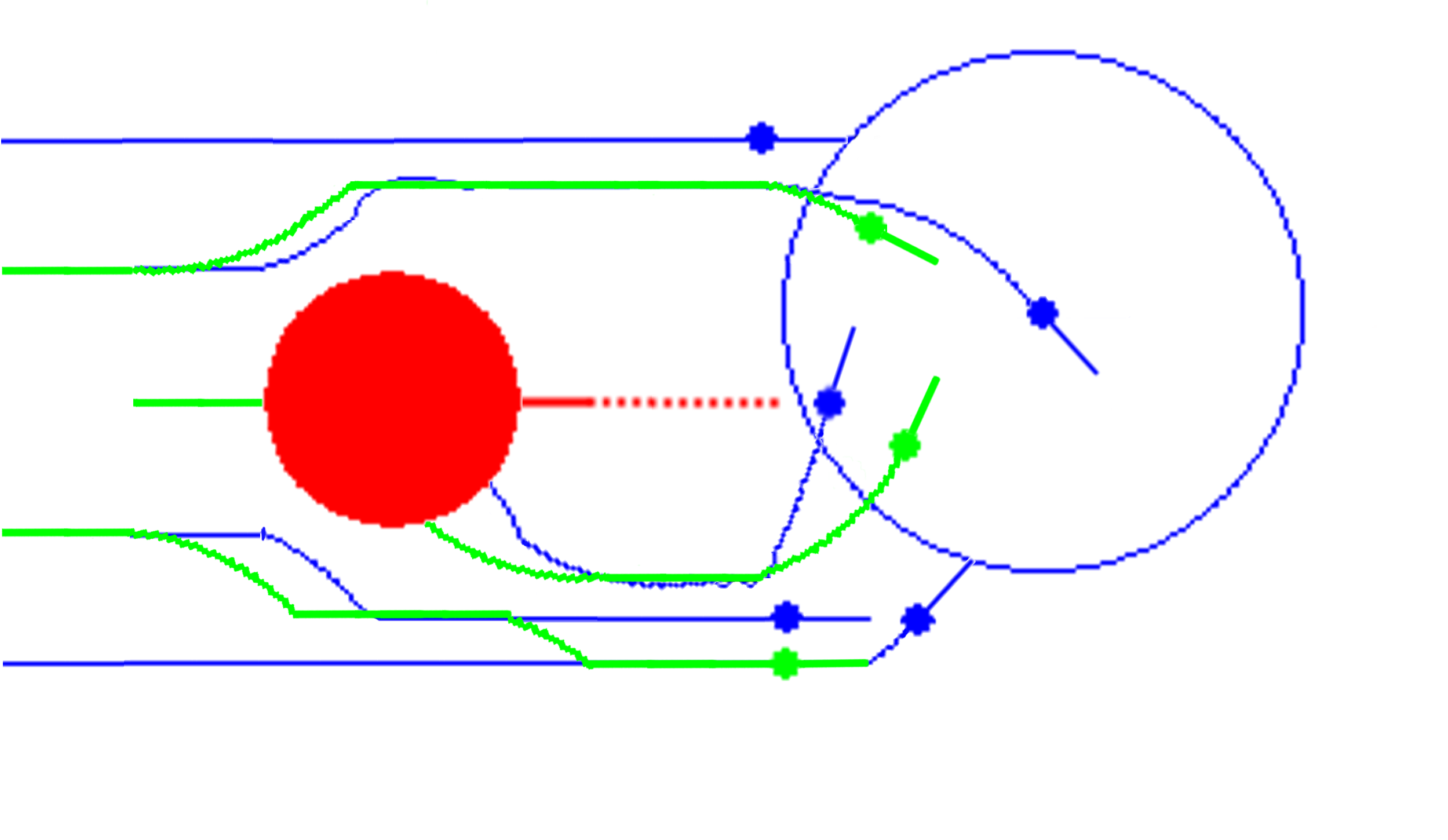}}
   \subfigure[\label{fig:init5}]{\includegraphics[width=0.35\columnwidth]{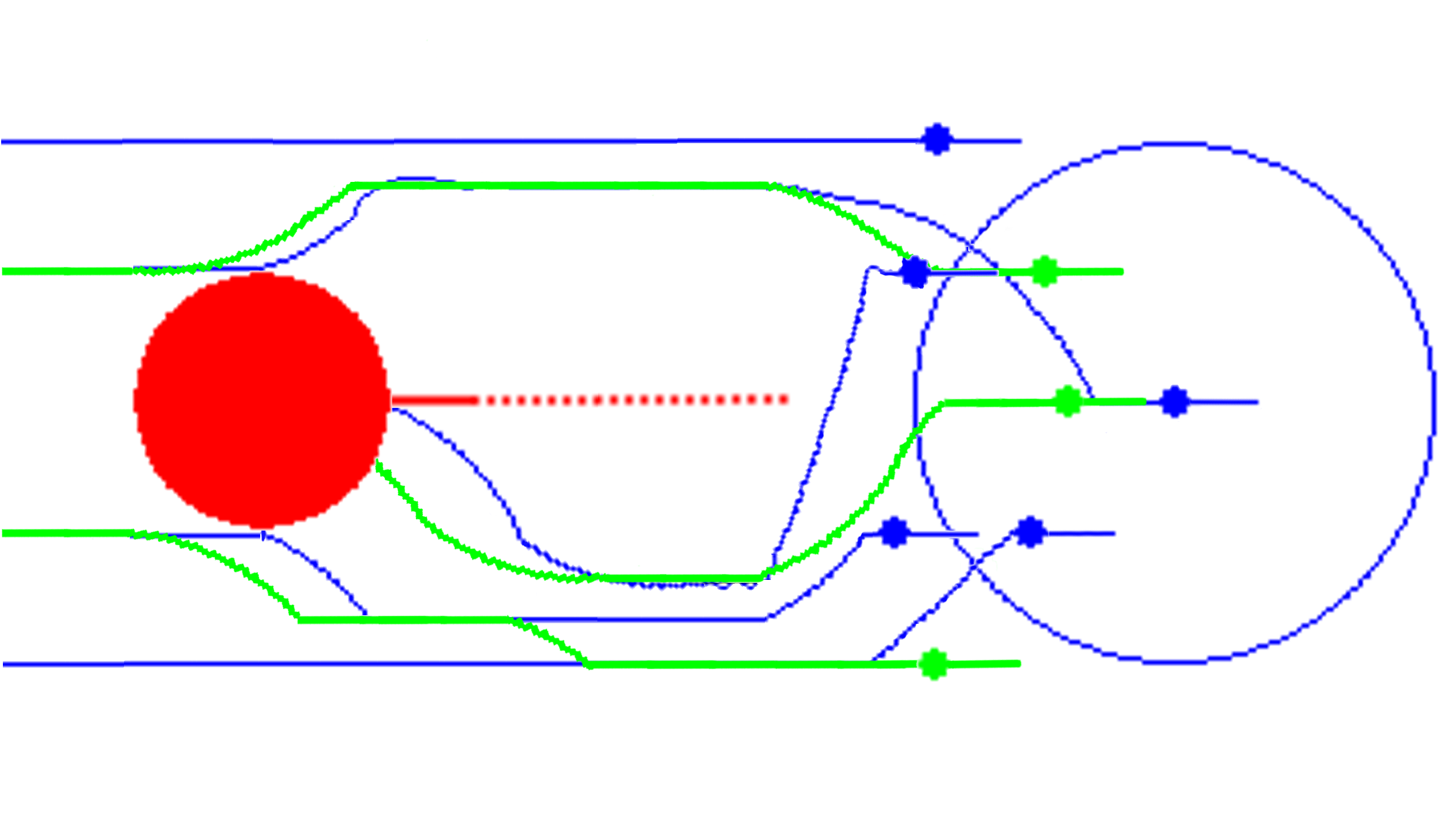}}
   \end{center}
  \caption{Simulation snapshots at equal time intervals, i.e., 0\%, 25\%, 50\%, 75\%, 100\% (a) When the UAVs are spawned and obstacle is moving towards the swarm. (b) obstacle is in detection range of the UAV, and Point of Impact is calculated and shown, UAV1 is deviating from its original path in order to avoid the Danger Zone. (c) Bypassing the obstacle. (d) notice the change of leader while bypassing the obstacle. (e) Leader changed while bypassing the obstacle.}
  \label{fig:ini_detect2}
\end{figure*}

\begin{figure}[!ht]
\begin{center}
    \subfigure[\label{fig:blahblah}]{\includegraphics[width=0.7\linewidth]{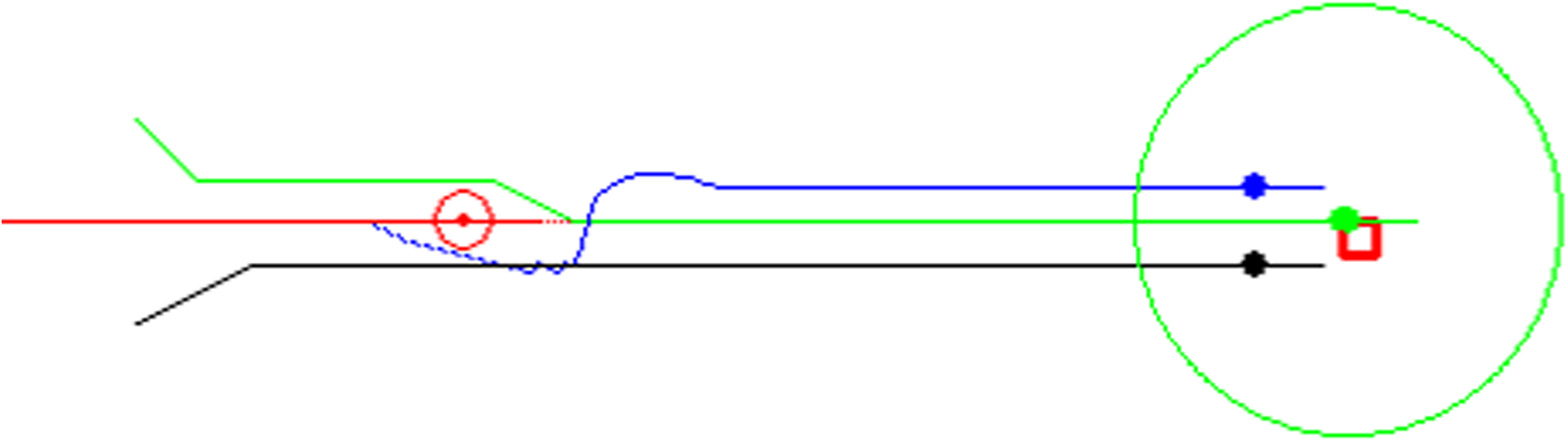}}
    \subfigure[\label{fig:blahblah2}]{\includegraphics[width=0.7\linewidth]{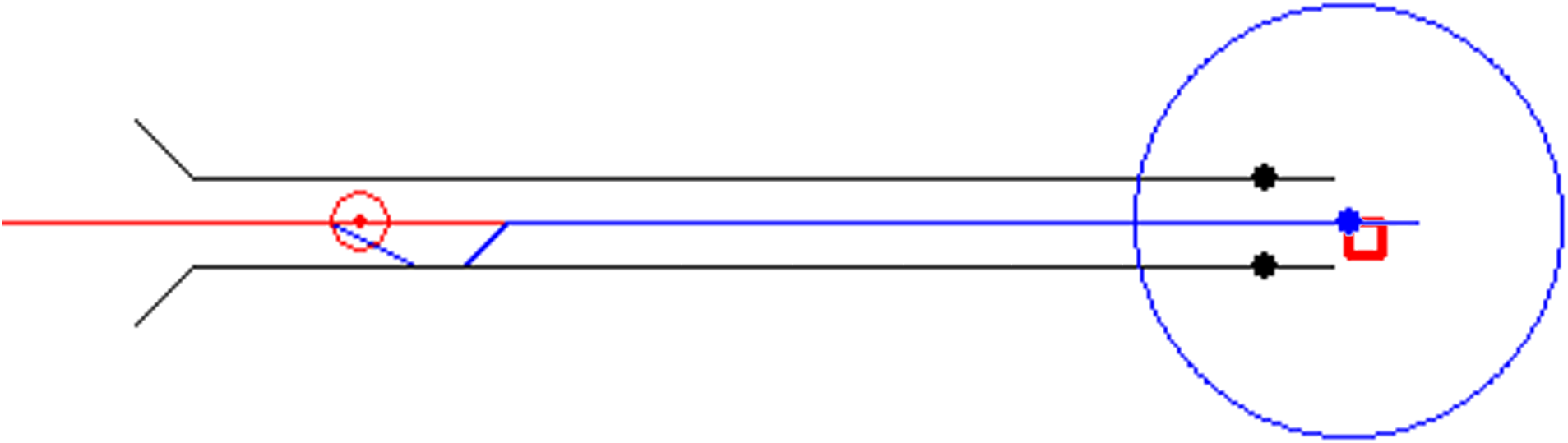}}
   \end{center}
  \caption{Swarm movement from start to destination. Navigational traces using: (a) proposed approach (b) dedicated leader}
  \label{fig:bvlahblah}
\end{figure}

The initial conditions/assumptions for our work are defined as follows:

\begin{enumerate}
    \item there is no explicit unique leader; the leader for the swarm changes dynamically according to the situation, i.e., the leadership is a temporary role
    \item UAVs accelerate or decelerate as needed.
    \item UAVs obtain their own position vectors using the on-board localization techniques.
    \item communication channel ideal, i.e., lossless
    \item an obstacle can be stationary or moving towards the swarm or away from the swarm with unknown velocity.
    \item for visualization purposes and to avoid the overlapping, the detection range circle of only the leader is shown.
\end{enumerate}

The UAVs are spawned at near the defined V-shaped formation (Figure \ref{fig:ini_detect}). The current leader UAV1 (blue), starts moving towards the destination and the other UAVs start moving towards their positions in the formation. An obstacle is also moving towards the swarm, but at this instant it is outside the detection range of the on-board sensor system of the UAVs, as shown in Figure \ref{fig:initial1}. In Figure \ref{fig:obs_detect}, the obstacle is already in the detection range and the Point of Impact and the Danger Zone has been computed, as explained in Algorithm \ref{algo2}.

Figure \ref{fig:bypasing}, shows the trend of escape route chosen by the collision avoidance module by deviating UAV1 to its right and slowing down the velocity of UAV3 to allow UAV1 stay on chosen route, as explained in Algorithm \ref{algo3}. Figure \ref{fig:uav2ahead} and \ref{fig:leaderchange} shows reformation process using CPSR, as shown that since UAV1 had to slow down and deviate to avoid the Danger Zone, UAV2 in the meantime is dynamically declared as the leader, as it continued its trajectory on the same path with same velocity, gets ahead of the rest instead of waiting for UAV1 to get back into its formation position. This is done to make sure time of arrival of centroid to the destination is minimized. Similarly, the optimal reformation would require UAV1 to go to UAV2's place and UAV3 would just speed up to catch with its position in the formation, as explained in Section II-C.

Figure \ref{fig:blahblah}, shows the movement of the swarm from starting point till it reaches the destination. In comparison, the behaviour of the swarm if point set registration is used with explicit unique leader and without balancing the centroid of the swarm is shown in Figure \ref{fig:blahblah2}.

\begin{figure}[!ht]
    \centering
    \subfigure[\label{fig:distandaxis}]{\includegraphics[width=0.49\columnwidth]{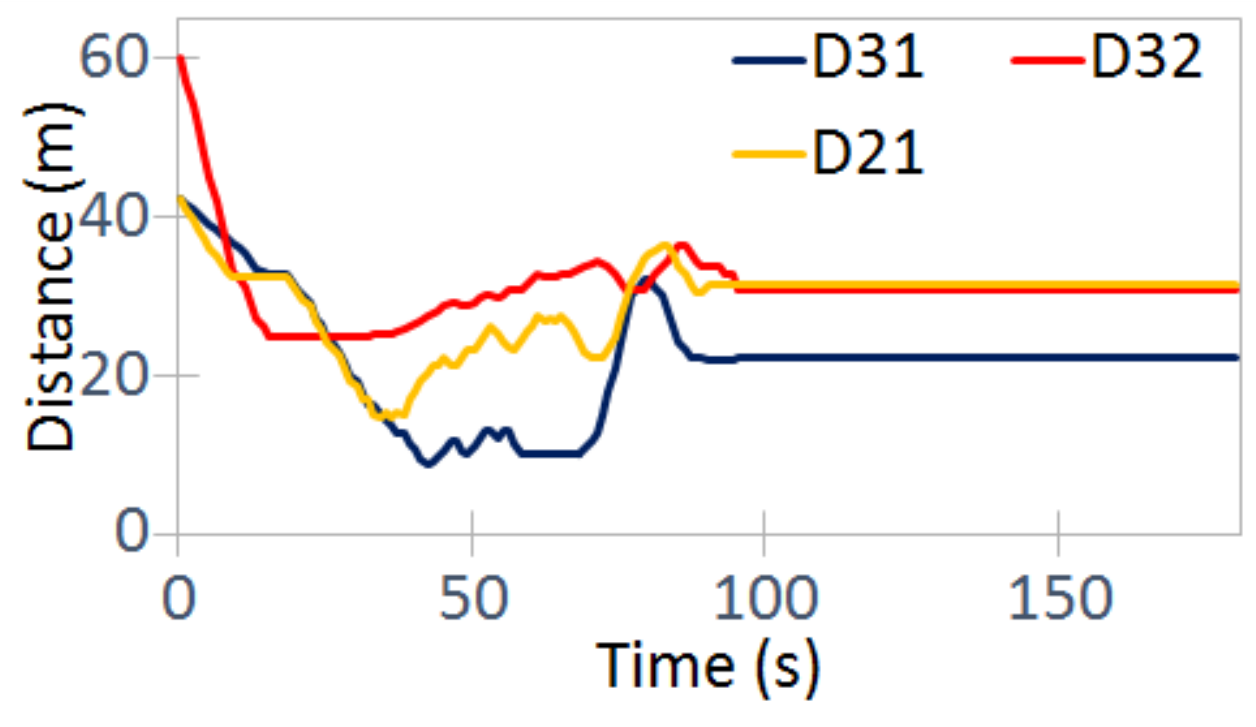}}
    \subfigure[\label{fig:chngtemp}]{\includegraphics[width=0.49\columnwidth]{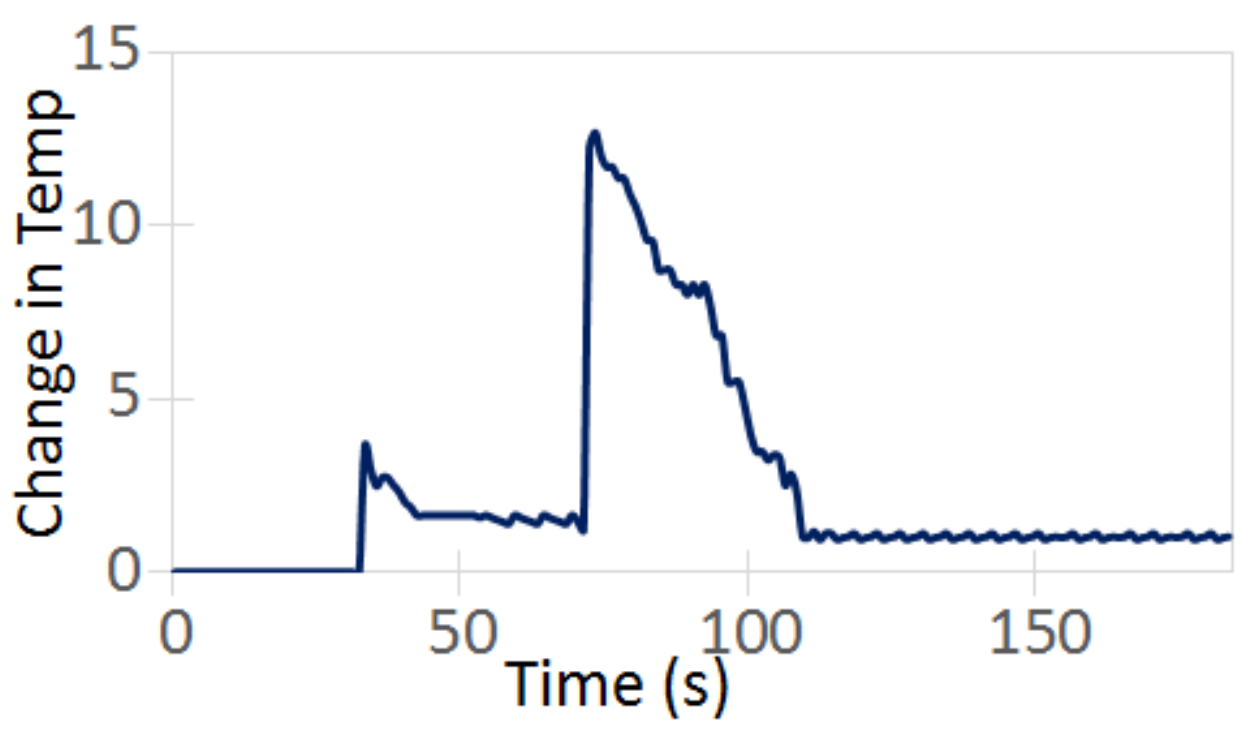}}
  \caption{(a) Distance maintained by drones with each other. (b) Change in Temperature of the system as a whole}
\end{figure}

The graph (Figure \ref{fig:distandaxis}) shows the overall trend of the distances maintained by the UAVs throughout the course, where D31 is the distance between UAV1 and UAV3, D21 is the distance between UAV1 and UAV2, and D32 is the distance between UAV2 and UAV3. The obstacle gets detected at $t = 30s$, and the collision avoidance is enforced which distorts the formation (Figure \ref{fig:ini_detect}).

In order to test the scalability of the proposed algorithm, the number of nodes in the swarm was increased to make a two layered V-shaped formation, as shown in Figure \ref{fig:ini_detect2}. The little overshoots in the movements of the drones in the figure can be reduced by integrating a more stable speed controller into the algorithm. Figure \ref{fig:chngtemp}, shows the change in the temperature of the system, i.e., the swarm, from the start until the destination is reached. At $t = 30s$, the disturbance/change in the temperature of the system shows the obstacle detection. The other significant disturbance at $t = 75s$ is due to the leader change and from there on the swarm gradually reshapes itself into the target shape defined by \textit{TShape}.

\begin{figure}[h]
    \centering
    \resizebox{0.8\columnwidth}{!}{
    \begin{tikzpicture}
    \begin{axis}[ height=3.5cm,width=6.9cm, x tick label style={
		 /pgf/number format/1000 sep=}, ybar, enlargelimits=.1, every node near coord/.append style={font=\tiny}, legend style={at={(0.5,-0.05)}, anchor=north,legend columns=-1}, symbolic x coords={}, ylabel= Time (s), ylabel near ticks, xtick=data, bar width=15pt, nodes near coords, nodes near coords align={vertical}, ymax=220 ]
        \addplot coordinates {(,166) };
    \addplot coordinates {(,181) };
    \addplot coordinates {(,198) };
    \addplot coordinates {(,213) };
    \legend{No obstacle,3-CPSR,8-CPSR,Unique}
    \end{axis} \end{tikzpicture}}
    \caption{Time for mission completion for different approaches\label{fig:bargraph}}
\end{figure}
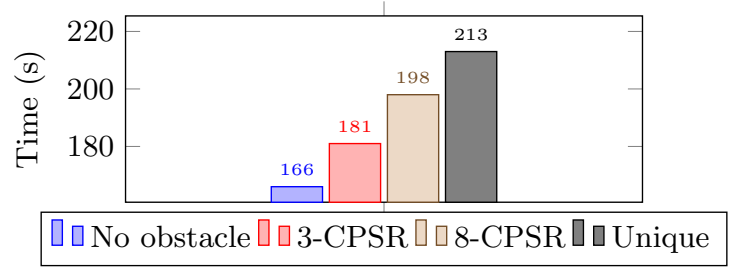

Figure \ref{fig:bargraph} shows the time taken for the swarm to reach its destination in four different scenarios: "No obstacle", i.e., if it is not disturbed at all and there are no deviations from the path, then $t_0 = 166s$; "Unique", i.e., if there is an obstacle in the path and the swarm has a unique dedicated leader, then $t_1 = 213s$; and finally two cases experimenting our proposed approach, i.e., in "3-CPSR" (a swarm with 3 nodes like in the previous cases) and "8-CPSR" (with 8 nodes), the time to finish was $t_2 = 181s$ and $t_3 = 198s$, respectively. This shows that we can considerably reduce the time by dynamically re-forming the swarm and changing the leader at run-time whenever the situation requires. The reason for this is that in the CPSR approach the swarm does not stop at any moment but keeps on progressing towards the destination, with each UAV deviating to avoid a collision when needed and accelerating afterwards to reach its position, defined by \textit{TShape}, to maintain the formation. On the other hand, in the fixed leader case ("Unique"), the swarm will slow down and wait for the leader to resume its position in the front before continuing towards the destination.

When considering the three-drone formation (Figure \ref{fig:ini_detect}), the swarm needed \textbf{\textit{56s}} from the obstacle detection to come back into the initial formation, whereas in the eight-drones case (Figure \ref{fig:ini_detect2}), this took \textbf{\textit{85s}}. It is evident from the experiments that in the latter case, due to a much bigger obstacle, the drones had to deviate more than in the former case. However, this did not affect the overall mission time very much, as UAV2, which became the new leader, did not have to deviate a lot from its path nor reduce its speed significantly.

\section{CONCLUSION}

In this paper, we developed a novel approach for collision avoidance and formation maintenance in a swarm of drones in dynamic environments. The proposed method utilizes a genetic algorithm inspired scheme in its collision avoidance part and point set registration in its formation maintenance part. In the approach, a swarm does not have a uniquely determined leader, and formation maintenance is accomplished by stabilizing the centroid of the swarm. The behaviour of the algorithm was theoretically analysed and tested in a simulation environment. The simulation results shown provide sufficient proof that the method works in a near-optimal manner in a dynamic environment, where an obstacle continues movement in its detected trajectory. We tested the efficiency of the proposed algorithm by comparing it with corresponding algorithms that assume existence of an explicit/unique leader. It was demonstrated that the ability to re-elect the leader dynamically, if required, gets the mission completed more quickly, i.e., it saves time and consequently energy by sparing the swarm from waiting for the leader to get back into its defined position in the formation.

In our future work, we plan to extend the proposed approach by examining the other environmental effects, such as air drag, on the layers of drones, such as a two or multi layered V-shaped formation. That can help in optimizing the resource management in the swarm by dynamically swapping the outer layer with the inner layer in order to minimize the effect of air drag and maximize the flight time on a single charge. Also, we will consider more complex scenarios with several simultaneous obstacles and more versatile movement of obstacles.

\section*{Acknowledgement}
This work has been supported in part by the Academy of Finland-funded research project 314048 and Nokia Foundation (Award No. 20200147).

%
%
\bibliographystyle{splncs03_unsrt}
\bibliography{template.bib}

\end{document}